\documentclass[11pt]{article}
\usepackage[margin=1in]{geometry}
\usepackage[T1]{fontenc}
\usepackage{times}
\usepackage{amsmath}
\usepackage{graphicx}
\usepackage{booktabs}
\usepackage[hidelinks]{hyperref}
\usepackage{xurl}
\usepackage{microtype}
\graphicspath{{figures/}{./}}

\hypersetup{
  pdftitle={Framing-Induced Vulnerability in Large Language Models: Evidence from Chinese News Headlines},
  pdfauthor={Mo Zhou and Yiwei Zhang}
}
\title{Framing-Induced Vulnerability in Large Language Models:\\ Evidence from Chinese News Headlines}
\author{Mo Zhou\\Nanjing Normal University \and Yiwei Zhang\\Communication University of China}
\date{}
\begin{document}
\maketitle
\begin{abstract}
Large language models increasingly evaluate, route, and moderate news, so a model's judgment of a headline has become a gatekeeping decision. We ask whether that judgment follows the fact being reported or the way the fact is framed. We combine an ecological audit of 63,926 Chinese news headlines with a controlled experiment. The experiment covers 30 real events, 30 synthetic events, and 29 natural headline pairs, tested on four Chinese models. Changing only the headline frame shifted both perceived misleadingness and event severity, and the direction and ordering of the shifts were the same for DeepSeek, MiMo, Dots, and LongCat. Emotionally intensified headlines raised severity ratings by 0.65 to 0.88 points, whereas positively reframed headlines lowered them by 0.34 to 0.52 points. Supplying the article body reduced this severity bias by 54 to 80 percent, and all four reductions were significant at $p<.01$. A small human reference study showed the same direction of effect. Headline-only LLM evaluation therefore carries a measurable robustness gap, and framing sensitivity belongs among the reliability properties that deployment and governance check.
\end{abstract}

\section{Introduction}
A headline can change the reading of a fact without changing the fact. Framing research established this property for human audiences. Selection and salience shape how a problem is defined, explained, evaluated, and acted on \cite{entman1993}, and logically equivalent descriptions of one outcome can reverse a stated preference \cite{tk1981}. The property now carries operational weight. Language models classify, summarize, rank, and moderate news before most readers reach it, which places the model at an evaluative step that feeds search, recommendation, fact-checking, and moderation pipelines \cite{wallace2018,nechushtai2019}.

Prior work has documented framing effects in human judgment, shown that model outputs move when source labels change \cite{germani2025}, and used models to detect misleading headlines \cite{rony2024}. Two gaps remain. Controlled model experiments are rarely tied to the framing practices of an actual news ecology, so it is unclear whether laboratory manipulations resemble what outlets publish. It is also unclear whether giving the model the article body removes the bias or merely dilutes it. We address both gaps with two linked studies. Study~1 audits the scale and distribution of framing intensity in Chinese online news. Study~2 tests whether matched headline variants change model judgments, and whether article context attenuates the change.

Our contributions are threefold. First, we provide an ecological corpus and a five-level framing annotation of 63,926 Chinese headlines. Second, we show that headline framing shifts misleadingness and severity judgments across four independently developed model families. Third, we show that supplying the article body lowers the severity bias substantially, which identifies a concrete mitigation for headline-only deployment.

\section{Related Work and Research Questions}
\subsection{Framing and model judgment}
Framing theory describes how selection and salience shape problem definition, causal interpretation, evaluation, and recommended action \cite{entman1993}. Equivalence framing changes the wording of the same information, and emphasis framing makes some aspects more salient than others. Headlines instantiate both forms through numerical presentation, evaluative verbs, emotional terms, and omitted denominators. Human readers respond to manipulative headlines less than the clickbait debate assumes. In a large survey experiment, clickbait exposure did not measurably raise polarization, distrust, or false recall \cite{munger2020}. Whether models show comparable sensitivity, or more, is an open question, and one that matters because models now label headlines at scale \cite{rony2024}.

\subsection{Evaluative gatekeeping}
Algorithmic gatekeeping research examines how systems select and rank information \cite{wallace2018,nechushtai2019}. We focus on an earlier point in the pipeline: evaluation. A model that labels a headline as misleading or severe can shape downstream decisions even when it neither writes nor distributes the content. We call this evaluative gatekeeping.

\subsection{Research questions}
\textbf{RQ1:} Does changing the frame of a headline about the same event systematically change LLM judgments of misleadingness and severity?\\[2pt]
\textbf{RQ2:} How do numerical presentation and evaluative wording contribute to severity judgments, and does article context reduce the resulting bias?

\section{Study 1: Ecological Audit}
We collected Chinese online news headlines published between January and September 2026 from Xinhua, China News Service, and Jingchu Net. After deduplication and validity filtering, 63,926 headlines were coded on five framing levels: L1 objective description, L2 objective description with directional wording, L3 emotional or numerical shock, L4 suspense or threat, and L5 simultaneous manipulation on several dimensions. Key dimensions were checked against a gold standard, with Cohen's $\kappa \geq .90$.

\begin{figure}[htbp]
\centering
\includegraphics[width=\textwidth]{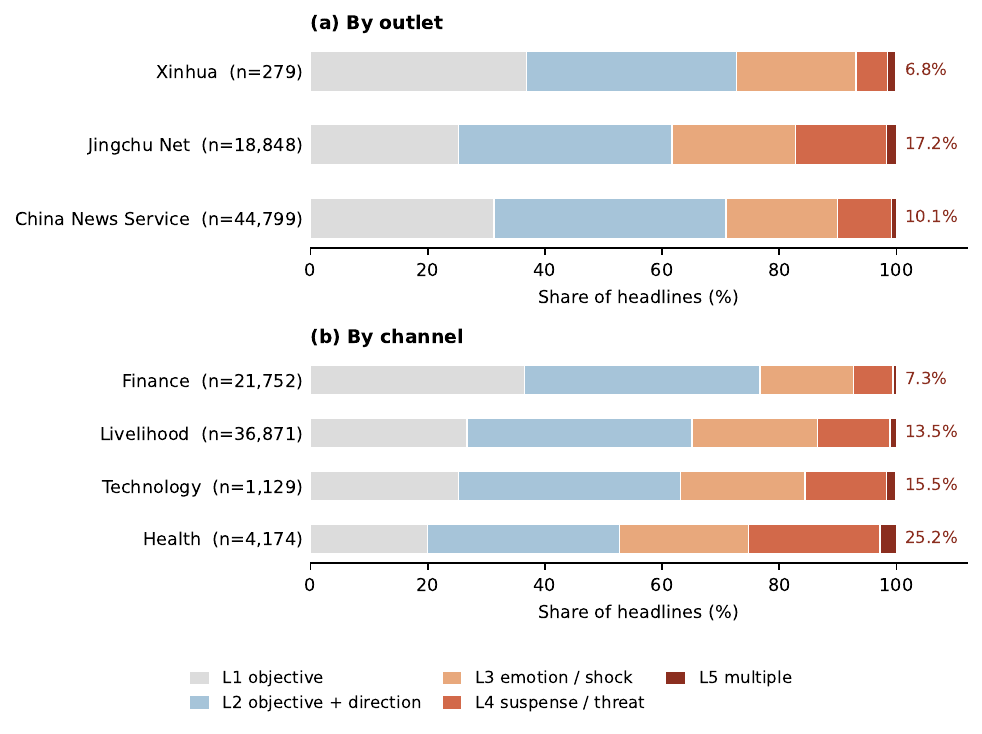}
\caption{Framing intensity in Chinese online news, by outlet (\textbf{a}) and by channel (\textbf{b}). Segments give the share of headlines at each framing level; the value at the right of each bar is the combined L4--L5 share. $n=63{,}926$ headlines collected between January and September 2026.}
\label{fig:intensity}
\end{figure}

\begin{figure}[htbp]
\centering
\includegraphics[width=0.8\textwidth]{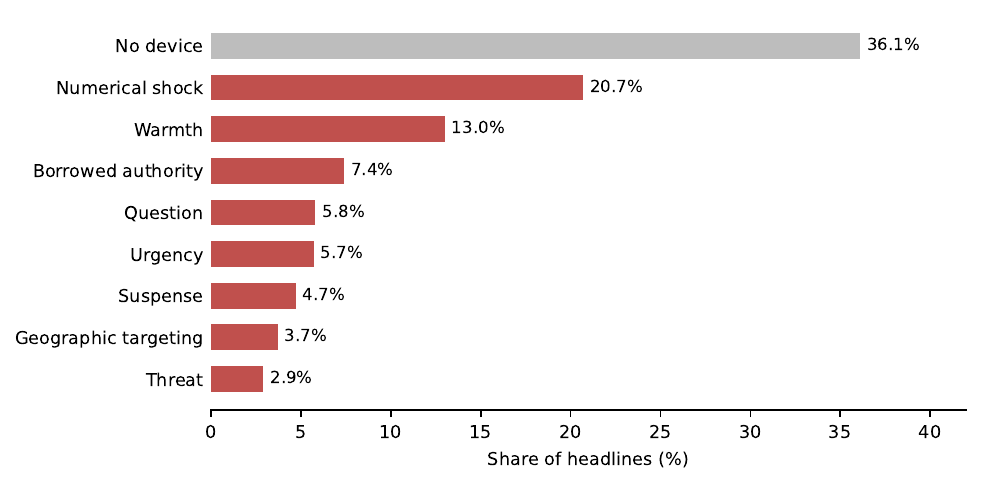}
\caption{Manipulative devices in the headline corpus. Shares are computed over all 63,926 coded headlines; 36.1\% carried no identifiable device and the remainder show the most frequent single devices.}
\label{fig:devices}
\end{figure}

Framing intensity varied by outlet and topic (Figure~\ref{fig:intensity}). High-intensity framing (L4--L5) accounted for 17.2\% of Jingchu Net headlines, 10.1\% of China News Service headlines, and 6.8\% of Xinhua headlines. Across channels, health carried the largest L4--L5 share (25.2\%), followed by technology (15.5\%), people's livelihood (13.5\%), and finance (7.3\%). Numerical shock was the most common manipulative device, present in 20.7\% of headlines (Figure~\ref{fig:devices}), which motivated the numerical and wording manipulations in Study~2.

\section{Study 2: Controlled Model Experiment}
\subsection{Materials and models}
\begin{figure}[htbp]
\centering
\includegraphics[width=\textwidth]{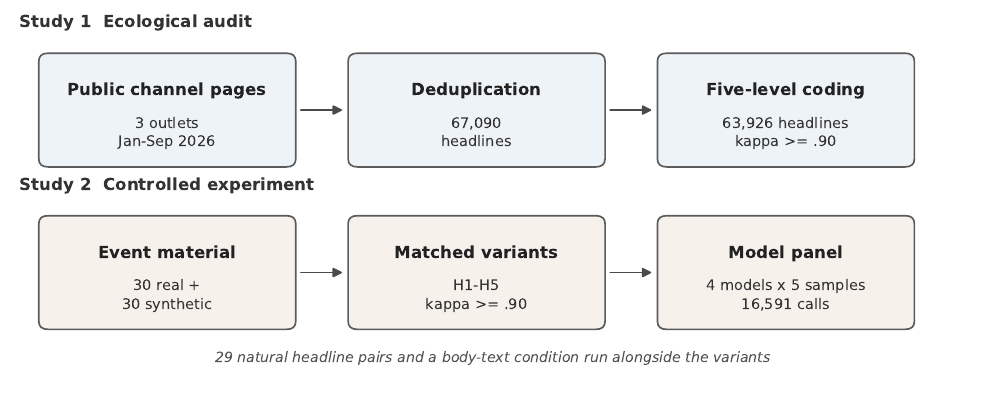}
\caption{Study design. Study~1 codes the ecological corpus; Study~2 builds matched variants for 60 events and evaluates them on a four-model panel. The natural pairs and the body-text condition run alongside the matched variants.}
\label{fig:pipeline}
\end{figure}

We built matched variants for 30 real events and 30 synthetic events, and each event had a fact card verified against authoritative sources. Five headline variants were written: H1 objective baseline, H2 negative verb, H3 emotional wording with an approximate number, H4 positive reframing, and H5 passive restatement. We also included 29 naturally occurring objective/manipulative headline pairs and a body-text condition for the same events. All variants passed a double-blind manipulation check ($\kappa \geq .90$). Figure~\ref{fig:pipeline} summarizes the design.

The model panel was DeepSeek-v4-flash, MiMo-v2.5-pro, Dots3-note-preview, and LongCat-2.0. Every headline-task combination was sampled five times at temperature 0.3, yielding 16,591 valid calls in the main experiment. A pre-experiment event-knowledge probe (900 calls across six models) found awareness of the underlying facts ranging from 7.3\% to 24.7\%. No awareness was recorded for 11 of the 30 real events. We therefore treat training-data contamination as a limit on interpretation rather than a threat that has been ruled out.

\subsection{Tasks and analysis}
Each trial asked the model to rate how misleading the headline was (T1) and how severe the event was (T2) on five-point scales. We report condition contrasts, model-wise ranges, and the reduction in bias achieved with article context. Trials are clustered within events, and the contrasts below are averaged differences rather than mixed-effects estimates; we return to this point in the limitations.

\section{Results}
\subsection{RQ1: misleadingness}
\begin{figure}[htbp]
\centering
\includegraphics[width=\textwidth]{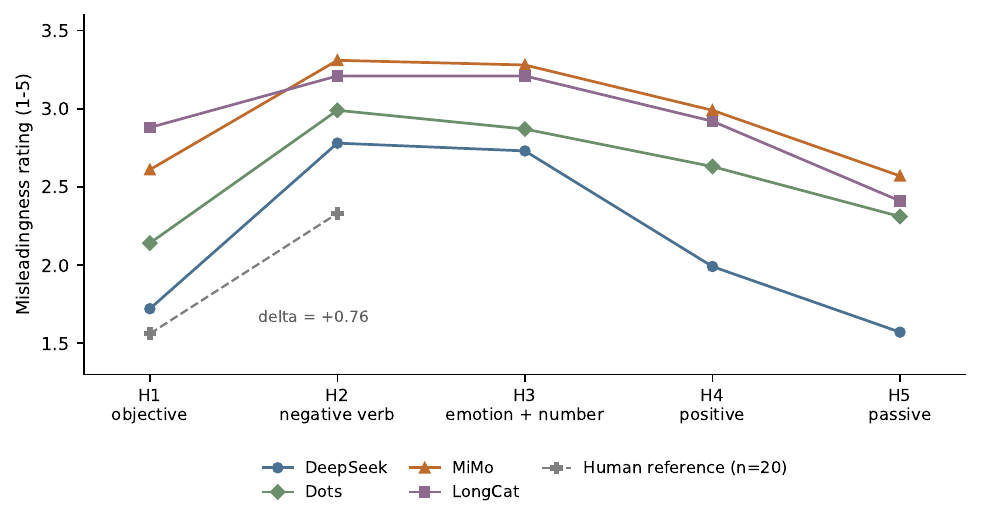}
\caption{Misleadingness ratings by headline variant. Lines give model means on the five-point scale. The dashed segment gives the human reference contrast between H1 and H2 for the 30 real events ($n=20$ raters, 600 ratings).}
\label{fig:profile}
\end{figure}

Changing only the headline frame raised misleadingness ratings. Relative to H1, the H2 contrast was +1.06 for DeepSeek, +0.85 for Dots, +0.70 for MiMo, and +0.33 for LongCat. The ordering H2 $\approx$ H3 $>$ H4 $>$ H1 $>$ H5 held for all four models (Figure~\ref{fig:profile}). In the 29 natural pairs, the manipulative headline received the higher misleadingness rating in every model. The real-event and synthetic-event tracks moved in parallel, which supports reading the effect as a property of presentation rather than of event realism. Table~\ref{tab:contrasts} collects the model-level contrasts.

\subsection{RQ2: severity}
H3 raised severity ratings by 0.65 to 0.88 points across models, and H4 lowered them by 0.34 to 0.52 points. The narrow cross-model range indicates a stable directional effect. Because H3 combines emotional wording with an approximate quantity, the result should be read as a joint framing effect rather than as a pure denominator manipulation.

\subsection{Article context}
\begin{figure}[htbp]
\centering
\includegraphics[width=0.9\textwidth]{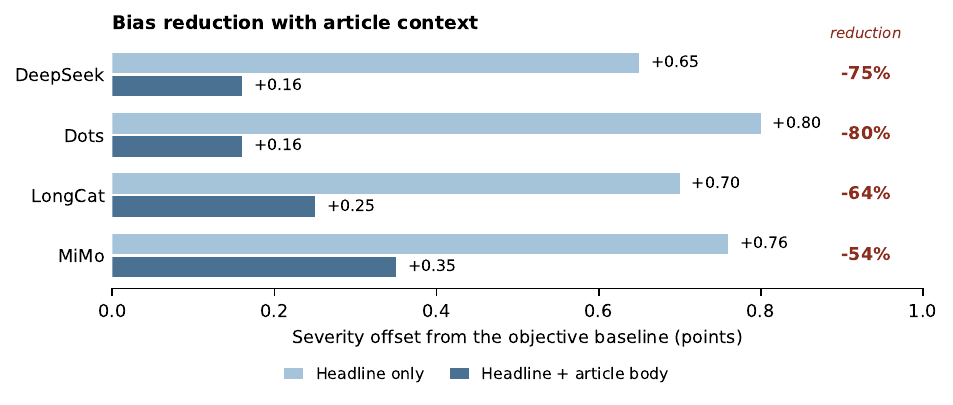}
\caption{Severity offset from the objective baseline, with and without the article body. Labels on the right give the bias reduction achieved by adding body text. All four reductions were significant at $p<.01$.}
\label{fig:reduction}
\end{figure}

When the article body was supplied with the headline, the severity bias fell by 54 to 80 percent across the four models: DeepSeek 75\%, Dots 80\%, LongCat 64\%, and MiMo 54\%. Each reduction was significant at $p<.01$. Context therefore attenuates the effect without removing it, because the residual bias stayed positive in every model (Figure~\ref{fig:reduction}).

\begin{table}[htbp]
\centering
\small
\caption{Model-level severity contrasts and natural-pair contrasts. Values are differences in points on the five-point scale. The two severity columns compare each variant with the objective baseline (T2); the natural-pair columns compare 29 manipulative headlines with their objective counterparts on misleadingness (T1) and severity (T2). Misleadingness by variant is shown in Figure~\ref{fig:profile} and is not repeated here.}
\label{tab:contrasts}
\begin{tabular}{lcccc}
\toprule
Model & $\Delta$H3$-$H1 & $\Delta$H4$-$H1 & Natural pairs & Natural pairs \\
 & (T2) & (T2) & (T1) & (T2) \\
\midrule
DeepSeek-v4-flash  & $+0.65$ & $-0.47$ & $+1.32$ & $+0.19$ \\
Dots3-note-preview & $+0.88$ & $-0.52$ & $+0.76$ & $+0.54$ \\
MiMo-v2.5-pro      & $+0.76$ & $-0.47$ & $+0.59$ & $+0.39$ \\
LongCat-2.0        & $+0.71$ & $-0.34$ & $+0.97$ & $+0.32$ \\
\bottomrule
\end{tabular}
\end{table}

\subsection{Human reference study}
Twenty participants rated real-event headlines in a planned-missingness design. Each participant saw 30 headlines, one variant per event, and each headline received ten ratings, giving 600 ratings in total. The event-level contrast between H2 and H1 was +0.76, $t(29)=3.74$, $p<.001$, $d_z=.68$. Model effects ran in the same direction. The sample is too small to support a claim of human--model equivalence.

\section{Discussion}
The results extend framing research from human judgment to model evaluation. Models that are treated as neutral evaluators track headline-level selection, salience, and wording, and they do so across four independently developed families. The exposure matters most where systems consume headlines without the article body, as in lightweight search, triage, and moderation pipelines. Supplying body context is a cheap mitigation, but it lowers the bias rather than removing it.

The study also names a governance target. Framing sensitivity should be measured alongside factuality and calibration. Deployment reports should state whether the model sees headlines only, whether source and body context are available, and how repeated stochastic outputs are aggregated.

\section{Limitations and Ethics}
The event set covers Chinese online news and a single 2026 model snapshot. The size and ordering of the effects are therefore not established for other languages or later model releases. Synthetic events give experimental control but do not establish external validity. Several headline variants change more than one linguistic dimension at a time, which limits causal attribution to a single feature. The human study is a directional reference rather than a powered comparison. Because trials are clustered within events, the reported contrasts carry no event-level random effects or confidence intervals, and a mixed-effects reanalysis may narrow or widen the model-wise ranges. The corpus was collected from public channel pages, and the human-validation data contain no participant-identifying information.

\section{Data Availability}
The cleaned corpus is available at \url{https://github.com/yadnuses/cn-news-headline-framing-dataset} under a CC BY-NC 4.0 license. It contains 67,090 deduplicated headlines with title, date, outlet, and channel; the analysis sample of 63,926 applies the additional validity filter described in Study~1. Source URLs, article bodies, and per-headline coding labels are withheld from the public release, and annotation guidelines and analysis code are available from the authors on request.

\section{Conclusion}
Headline framing systematically changes LLM judgments of misleadingness and severity across four Chinese models. Article context reduces the severity bias by 54 to 80 percent, which sets both a concrete mitigation and a deployment boundary. Robust headline evaluation should therefore treat presentation sensitivity as a first-class reliability dimension.

\end{document}